\documentclass[runningheads]{llncs}
\usepackage{graphicx}

\usepackage{tikz}
\usepackage{comment}
\usepackage{amsmath,amssymb} 
\usepackage{color}

\usepackage[accsupp]{axessibility}  

\usepackage{subcaption}
\usepackage{colortbl}
\usepackage{tabulary}
\usepackage{etoolbox}
\usepackage{booktabs}

\usepackage{enumitem}
\usepackage{xcolor}

\definecolor{ForestGreen}{rgb}{0.13, 0.55, 0.13}
\usepackage{pifont}
\newcommand{\cmark}{{\color{ForestGreen} \ding{51}}}%
\newcommand{\xmark}{{\color{red} \ding{55}}}%

\renewcommand{\paragraph}[1]{\textbf{#1.}}

\usepackage{cuted}

\usepackage{adjustbox}
\usepackage{array}

\newcolumntype{R}[2]{%
    >{\adjustbox{angle=#1,lap=\width-(#2)}\bgroup}%
    l%
    <{\egroup}%
}
\newcommand*\rot{\multicolumn{1}{R{60}{2em}}}

\usepackage{multirow}

\begin{document}
\pagestyle{headings}
\mainmatter
\def\ECCVSubNumber{2459}  

\title{Unsupervised Deep Multi-Shape Matching} 

\author{Dongliang Cao\inst{1,2} \and
Florian Bernard\inst{1}}
\authorrunning{D.~Cao and F.~Bernard}

\institute{University of Bonn, Germany \and Technical University of Munich, Germany}
\maketitle

\begin{abstract}
 3D shape matching is a long-standing problem in computer vision and computer graphics. While deep neural networks were shown to lead to state-of-the-art results in shape matching, existing learning-based approaches are limited in the context of multi-shape matching: (i) either they focus on  matching pairs of shapes only and thus suffer from cycle-inconsistent multi-matchings, or (ii) they require an explicit template shape to address the matching of a collection of shapes. In this paper, we present a novel approach for deep multi-shape matching that ensures cycle-consistent multi-matchings while not depending on an explicit template shape. To this end, we utilise a shape-to-universe multi-matching representation that we combine with powerful functional map regularisation, so that our multi-shape matching neural network can be trained in a fully unsupervised manner. While the functional map regularisation is only considered during training time, functional maps are not computed for predicting correspondences, thereby allowing for fast inference. We demonstrate that our method achieves state-of-the-art results on several challenging benchmark datasets, and, most remarkably, that our unsupervised method even outperforms recent supervised methods. 
\end{abstract}

\section{Introduction}
The matching of 3D shapes is a long-standing problem in computer vision and computer graphics.
Due to its wide range of applications, numerous approaches that address diverse variants of shape matching problems have been proposed over the past decades~\cite{van2011survey,tam2012registration}.
In recent years, with the success of deep learning, many learning-based methods were introduced for shape matching. One common way to address shape matching is to formulate it as classification
problem~\cite{masci2015geodesic,boscaini2016learning,monti2017geometric,fey2018splinecnn,li2020shape,wiersma2020cnns}. The advantage of such methods is that after training the classifier, shape correspondences can efficiently and directly be predicted. A major downside is that for training such a classifier typically a large amount of  data that is annotated with ground truth correspondences is required. 
However, specifically in the domain of 3D shapes, annotated data is scarce, since data annotation is particularly time-consuming and tedious.
Thus, in practice, the above methods are often trained with small datasets, so that in turn they are prone to overfitting and lack the ability to generalise across datasets. 

Another line of learning-based shape matching solutions build upon the functional map framework~\cite{litany2017deep,halimi2019unsupervised,roufosse2019unsupervised,donati2020deep,sharma2020weakly}. Functional maps can serve as a powerful regularisation, so that respective methods were even trained successfully in an unsupervised manner, i.e.~without the availability of ground truth correspondences~\cite{roufosse2019unsupervised,sharma2020weakly,attaiki2021dpfm}. 
However, a downside of such approaches is that the conversion of the obtained functional map to a point-wise correspondence map is typically non-trivial. On the one hand, this may limit the matching accuracy, while on the other hand this may have a negative impact on inference speed (see Sec.~\ref{sec:full_results}).

\newcommand{\teaserheight}[0]{2.75cm}
\begin{figure}[t!]
\setlength{\tabcolsep}{-2.5pt}
  \centerline{
  \scriptsize
  \begin{tabular}{ccc}
          \includegraphics[height=\teaserheight]{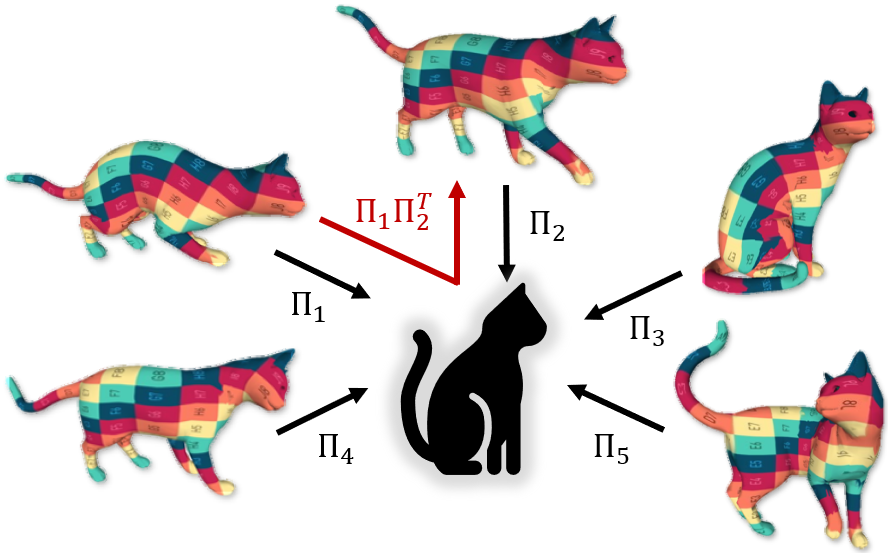} ~&~
          \includegraphics[height=\teaserheight]{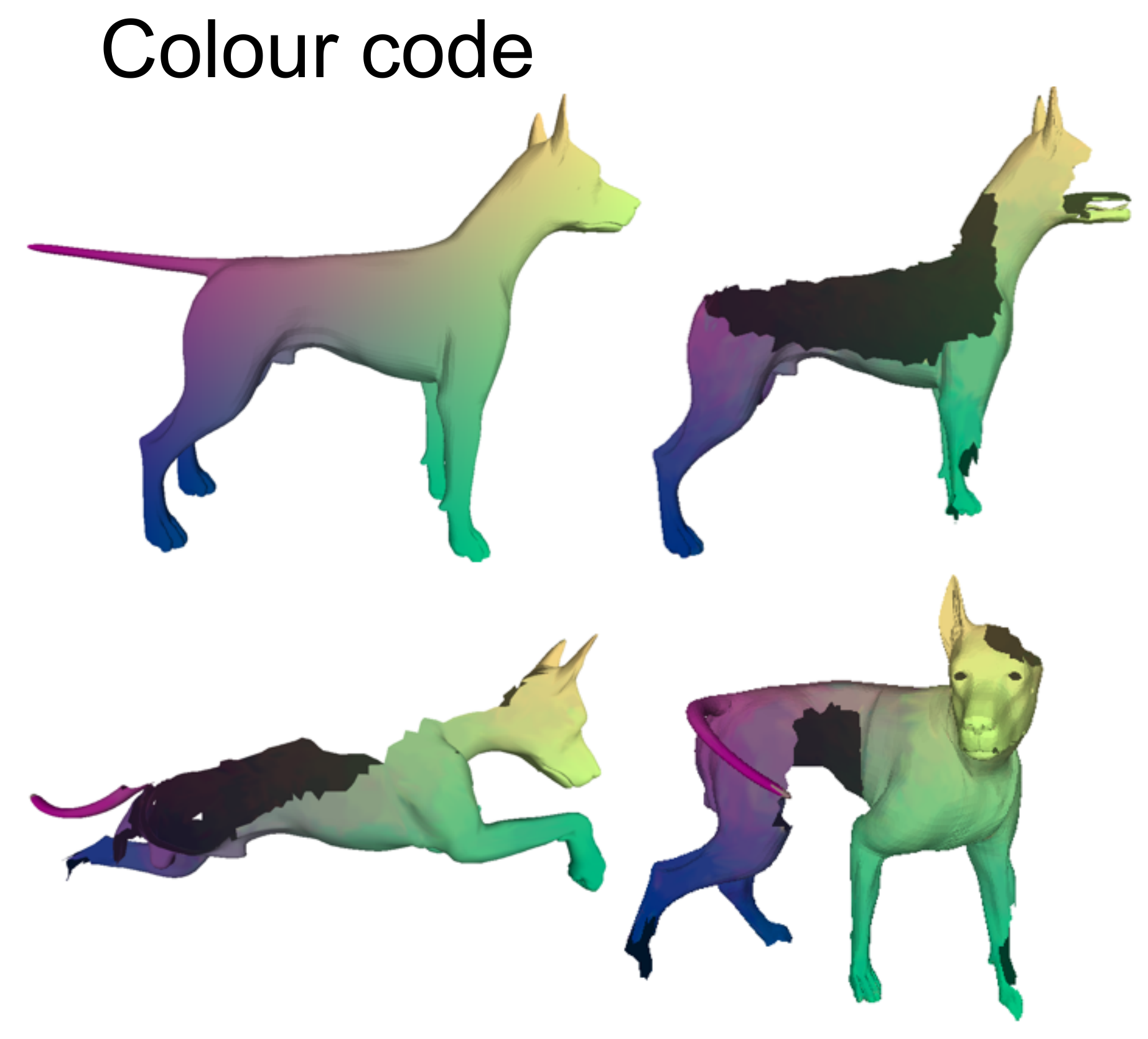} ~&
\resizebox{4cm}{!}{
\begin{tabular}[b]{llllllllll}
&\rot{FMNet~\cite{litany2017deep}}&\rot{SURFMNet~\cite{roufosse2019unsupervised}} & \rot{GeomFmaps~\cite{donati2020deep}}  & \rot{WSupFMNet~\cite{sharma2020weakly}}& \rot{DPFM~\cite{attaiki2021dpfm} }     & \rot{Deep Shells~\cite{eisenberger2020deep}} & \rot{3D-CODED~\cite{groueix20183d}}    & \rot{ACSCNN~\cite{li2020shape}}& \rot{\textbf{Ours}}\\ \hline
Unsup.          & \xmark  & \cmark  & \xmark & \cmark & \cmark & \cmark  & \cmark  & \xmark  & \cmark \\
Cycle-cons.      & \xmark  & \xmark  & \xmark & \xmark & \xmark & \xmark  & \cmark  & \cmark  & \cmark \\
No template~~  & \cmark  & \cmark  & \cmark & \cmark & \cmark & \cmark  & \xmark  & \xmark  & \cmark \\
Partiality     & \xmark  & \xmark  & \xmark & \cmark & \cmark & \xmark  & \xmark  & \xmark  & \cmark \\ \hline    \\  
\end{tabular}
} ~~ 
\\
          Our multi-matching representation & {Partial multi-matching} & Comparison of methods
        \end{tabular}
  }
    \caption{\textbf{Left}: We present a novel unsupervised learning approach for cycle-consistent multi-shape matching based on matching each shape to a (virtual) universe shape. 
    \textbf{Middle}: Our approach can successfully solve challenging \textit{partial} multi-shape matching problems. 
    \textbf{Right}: Our method is the first learning-based multi-shape matching approach that combines several favourable properties, i.e.~it can be trained in an unsupervised manner, obtains cycle-consistent multi-matchings, does not require a template, and allows for partial matchings.
    }
    \label{fig:teaser}
\end{figure}

In this work we aim to alleviate the shortcomings of both paradigms, while combining their advantages.
To this end,
we present a novel unsupervised learning approach for obtaining cycle-consistent multi-shape matchings, see Fig.~\ref{fig:teaser}. 
Our approach predicts shape-to-universe matchings based on  a \emph{universe classifier}, so that cycle-consistency of so-obtained pairwise matchings is guaranteed by construction.
Unlike previous classification methods that rely on supervision based on ground truth correspondences, the training of our universe classifier purely relies on functional map regularisation, thereby allowing for a fully unsupervised training. Yet, at inference time, our method does not require to compute functional maps, and directly predicts shape-to-universe matchings via our classifier.
We summarise our main contributions as follows:
\begin{itemize}
       \item For the first time we enable the \emph{unsupervised training} of a classification-based neural network for \emph{cycle-consistent 3D multi-shape matching}. 
       \item To this end, our method uses \emph{functional maps as strong regularisation during training}  but does \emph{not require the computation of functional maps during inference}. 
       \item Our method achieves \emph{state-of-the-art results} on several challenging 3D shape matching benchmark datasets, even in comparison to most recent supervised methods.
\end{itemize}

\section{Related work}
\label{sec: related_work}
Shape matching is a well-studied problem in computer vision and graphics~\cite{van2011survey,tam2012registration}. Rather than providing an exhaustive literature survey, in the following we will focus on reviewing those methods that we consider most relevant to our work.
First, we will provide an overview of works that rely on the functional map framework for 3D shape matching. Subsequently, we will focus on recent learning-based approaches that utilise functional maps. Afterwards, we will briefly discuss other learning-based methods.
Eventually, we will discuss methods that are specifically tailored for the case of multi-shape matching, as opposed to the more commonly studied case of two-shape matching.

\paragraph{Functional maps}
The functional map framework~\cite{ovsjanikov2012functional} enables to formulate the correspondence problem in the functional domain by computing functional maps instead of point-wise maps. The key advantage of functional maps is that they allow for a low-dimensional representation of shape correspondences, i.e.~small matrices that encode how functions can be transferred between two domains (3D shapes).
Unlike finding point-to-point correspondences, which can for example be phrased as the NP-hard quadratic assignment problem~\cite{lawler1963quadratic}, functional maps can efficiently be obtained by solving a linear least-squares problem. The functional map framework has been extensively studied and was extended in numerous works, e.g.~in terms of improving the accuracy or robustness~\cite{ren2019structured}, as well as extending its application to partial shape matching~\cite{rodola2017partial}, non-isometric shape matching~\cite{nogneng2017informative,ren2018continuous} and multi-shape matching~\cite{huang2014functional,huang2020consistent}. Nevertheless, in these approaches functional maps are typically utilised in an axiomatic manner, meaning that they heavily rely on handcrafted feature descriptors, such as HKS~\cite{bronstein2010scale}, WKS~\cite{aubry2011wave} or SHOT~\cite{salti2014shot}, which potentially limits their performance.

\paragraph{Learning methods based on functional maps}
In contrast to axiomatic approaches that use handcrafted features, a variety of methods have been proposed to learn the feature descriptors directly from data. Starting from~\cite{litany2017deep}, the (supervised) FMNet was proposed to learn a non-linear transformation of SHOT descriptors~\cite{salti2014shot}. Later work~\cite{halimi2019unsupervised} modified the loss to enable FMNet training in an unsupervised manner. However, both methods compute a loss that relies on geodesic distances, {which is computationally expensive, particularly for high-resolution shapes}. SURFMNet~\cite{roufosse2019unsupervised} proposed an unsupervised loss based on functional map regularisation, {which, however, does not directly obtain point-to-point correspondences.}
More recently, several works~\cite{donati2020deep,sharma2020weakly} replaced FMNet with point-based networks \cite{qi2017pointnet++,thomas2019kpconv} to achieve better performance. However, such point-based networks cannot utilise the connectivity information that exists in triangle meshes. To make use of it, DiffusionNet~\cite{sharp2020diffusionnet} introduced a diffusion layer, which was shown to achieve state-of-the-art performance for 3D shape matching. Most recently, DPFM \cite{attaiki2021dpfm} extended DiffusionNet with a cross-attention refinement mechanism~\cite{vaswani2017attention} for partial shape matching. 
DPFM addresses two variants of partial matching problems: for partial-to-partial matching it relies on a supervised training strategy, and for partial-to-complete matching it can be trained in an unsupervised manner based on functional map regularisation~\cite{rodola2017partial,litany2017fully}. While DPFM predicts functional maps and thereby requires a post-processing to obtain point-wise maps, our proposed approach directly predicts point-wise maps without the need of computing functional maps during inference.

\paragraph{Other learning-based methods}
Despite the success of functional maps for diverse learning-based shape matching approaches,
there is a wide variety of other learning-based methods. Many of the recent works formulate  shape matching as a classification problem~\cite{masci2015geodesic,boscaini2016learning,monti2017geometric,fey2018splinecnn,li2020shape,wiersma2020cnns}. In these methods, special network architectures were proposed to extend and generalise convolutions from Euclidean grids to surfaces. However, these methods require ground truth correspondences as supervision for training the classifier. 3D-CODED \cite{groueix20183d} factorised a given input shape into a template shape and a learned global feature vector that encodes the deformation from the template shape to the input shape. In this way, it finds correspondences between input shapes and the template shape. {In contrast, our method does not require to choose a template shape for matching}. Deep shells~\cite{eisenberger2020deep} proposed a coarse-to-fine matching pipeline that utilises an iterative alignment of smooth shells~\cite{eisenberger2020smooth}. While this iterative process may be time-consuming, our method directly predicts shape correspondences in one shot.

\paragraph{Multi-shape matching}
There are several works that explicitly consider the matching of a collection of shapes, i.e.~the so-called \textit{multi-shape matching problem}.
The key aspect in which multi-shape matching differs from the more commonly studied two-shape matching problem is that in the former one needs to ensure cycle-consistency between pairwise matchings across a collection of data. This is often achieved by first solving a quadratic number of pairwise matching problems (e.g.~between all pairs of shapes), and subsequently establishing cycle consistency as a post-processing, e.g.~based on permutation synchronisation~\cite{pachauri2013solving,huang2013consistent}.
The higher-order projected power iteration (HiPPI) method~\cite{bernard2019hippi} circumvented this two-stage approach by generalising permutation synchronisation to explicitly consider geometric relations.
In the context of functional maps, Consistent ZoomOut~\cite{huang2020consistent} extended ZoomOut~\cite{melzi2019zoomout} by adding functional map consistency constraints. IsoMuSh~\cite{gao2021isometric} simultaneously optimises for functional maps and point-wise maps that are both multi-shape consistent.
In contrast to these axiomatic methods, which usually
downsample the original shape for matching, our method
can be directly applied on shapes up to 10,000 vertices. 
Several learning-based approaches, such as 3D-CODED \cite{groueix20183d}, HSN~\cite{wiersma2020cnns} or ACSCNN~\cite{li2020shape}, utilised an explicit template shape in order to ensure cycle-consistent multi-matchings. However, the use of an explicit template shape poses severe limitations in practice, since a suitable template shape needs to be available, and the specific choice of template also induces a bias.
In stark contrast, our method does not rely on an explicit template shape, thereby effectively alleviating such a bias while at the same time providing substantially more flexibility.

\section{Background}
Our approach is based on the functional map framework and aims for a cycle-consistent multi-shape matching. For completeness, we first recap the basic pipeline for functional map computation and the desirable properties of functional maps. Then, we introduce the notion of cycle consistency for multi-shape matching.

\subsection{Functional maps}
\label{sec: fmaps}
\paragraph{Basic pipeline} Given is a pair of 3D shapes $\mathcal{X}$ and $\mathcal{Y}$ that are represented as triangle meshes with $n_x$ and $n_y$ vertices, respectively. The basic pipeline for computing a functional map between both shapes mainly consists of the following steps:
    \begin{itemize}[noitemsep,nolistsep]
        \item Compute the first $k$  eigenfunctions $\Phi_{x} \in \mathbb{R}^{n_{x} \times k}, \Phi_{y} \in \mathbb{R}^{n_{y} \times k}$ of the respective Laplacian matrix \cite{pinkall1993computing} as the basis functions.
        \item Compute feature descriptors $\mathcal{F}_{x} \in \mathbb{R}^{n_{x} \times c}, \mathcal{F}_{y} \in \mathbb{R}^{n_{y} \times c}$ on each shape, and (approximately) represent them in the (reduced) basis of the respective eigenfunctions, i.e.~$A_{x}=\Phi_{x}^{\dagger}\mathcal{F}_{x}, A_{y}=\Phi_{y}^{\dagger}\mathcal{F}_{y}$.
        \item Compute the optimal functional map $C_{xy} \in \mathbb{R}^{k \times k}$ by solving the optimisation problem 
        \begin{equation}
            \label{eq:fmap}
            C_{xy}=\underset{C}{\arg \min }~ \mathcal{L}_{\mathrm{data}}\left(C\right)+\lambda \mathcal{L}_{\mathrm{reg}}\left(C\right),
        \end{equation}
        where the data term $\mathcal{L}_{\mathrm{data}}$ ensures that $C$ maps between the feature descriptors represented in the reduced basis, 
        and the regularisation term $\mathcal{L}_{\mathrm{reg}}$ penalises the map by its structural properties (as explained below).
        \item Convert the functional map $C_{xy}$ to a point map $\Pi_{yx} \in \{0, 1\}^{n_{y} \times n_{x}}$, e.g.~using nearest neighbour search or other post-processing techniques~\cite{melzi2019zoomout,pai2021fast,vestner2017product} based on the relationship 
        \begin{equation}
            \label{eq:relation}
            \Phi_{y}C_{xy} \approx \Pi_{yx}\Phi_{x}.
        \end{equation}
    \end{itemize}

\noindent
\paragraph{Structural properties} In the context of near-isometric shape pairs, functional maps have the following properties~\cite{roufosse2019unsupervised,sharma2020weakly}:
    \begin{itemize}[noitemsep,nolistsep]
        \item \textbf{Bijectivity.} Given functional maps in both directions $C_{xy}, C_{yx}$, bijectivity requires the map from $\mathcal{X}$ through $\mathcal{Y}$  to $\mathcal{X}$ to be the identity. The requirement can be formulated as the difference between their composition and the identity map \cite{eynard2016coupled}. Thus, the bijectivity regularisation for functional maps can be expressed in the form
        \begin{equation}
            \label{eq:bij}
            \mathcal{L}_{\mathrm{bij}}=\left\|C_{xy} C_{yx}-I\right\|^{2}_{F}+\left\|C_{yx} C_{xy}-I\right\|^{2}_{F}.
        \end{equation}
        \item \textbf{Orthogonality.} A point map is locally area-preserving if and only if the associated functional map is an orthogonal matrix \cite{ovsjanikov2012functional}. The orthogonality regularisation for functional maps can be expressed in the form
        \begin{equation}
            \label{eq:orth}
            \mathcal{L}_{\mathrm{orth}}=\left\|C_{xy}^{\top} C_{xy}-I\right\|^{2}_{F}+\left\|C_{yx}^{\top} C_{yx}-I\right\|^{2}_{F}.
        \end{equation}
        \item \textbf{Laplacian commutativity.} A point map is an intrinsic isometry if and only if the associated functional map commutes with the Laplace-Beltrami operator \cite{ovsjanikov2012functional}. The Laplacian commutativity regularisation for functional maps can expressed in the form
        \begin{equation}
            \label{eq:lap}
            \mathcal{L}_{\mathrm{lap}}=\left\|C_{xy} \Lambda_{x}-\Lambda_{y} C_{xy}\right\|^{2}_{F}+\left\|C_{yx} \Lambda_{y}-\Lambda_{x} C_{yx}\right\|^{2}_{F},
        \end{equation}
        where $\Lambda_{x}$ and $\Lambda_{y}$ are diagonal matrices of the Laplace-Beltrami eigenvalues on the respective shapes.
    \end{itemize}

\subsection{Multi-shape matching}
Given is a collection 3D shapes $\mathcal{S}$. For any pair $\mathcal{X}, \mathcal{Y} \in \mathcal{S}$ with $n_{x}, n_{y}$ vertices, respectively, the point map $\Pi_{xy}$ between them can be expressed in the form
    \begin{equation}
        \label{eq:pmap}
        \Pi_{xy} \in \left\{\Pi \in\{0,1\}^{n_{x} \times n_{y}}: \Pi \mathbf{1}_{n_{y}} \leq \mathbf{1}_{n_{x}}, \mathbf{1}_{n_{x}}^{\top} \Pi \leq \mathbf{1}_{n_{y}}^{\top}\right\},
    \end{equation}
where $\Pi_{xy}(i, j) = 1$ can be interpreted as the $i$-th vertex in shape $\mathcal{X}$ corresponding to the $j$-th vertex in shape $\mathcal{Y}$. 

Cycle-consistency is a desirable property between pairwise matchings in a collection, as it must hold for the true matchings. Cycle consistency means that for any given triplet $\mathcal{X}, \mathcal{Y}, \mathcal{Z} \in \mathcal{S}$, the matching composition from $\mathcal{X}$ through $\mathcal{Y}$ to $\mathcal{Z}$ should be identical to the direct matching from $\mathcal{X}$ to $\mathcal{Z}$, i.e. 
    \begin{equation}
        \label{eq: cycle}
        \Pi_{xz} = \Pi_{xy}\Pi_{yz}.
    \end{equation}
We note that if cycle consistency holds for all triplets in $\mathcal{S}$, it also holds for any higher-order tuples of matching compositions, since the latter can be constructed by composing triplet matching compositions.

Alternatively, one can use a shape-to-universe matching representation \cite{pachauri2013solving,huang2013consistent} to avoid explicitly modelling the (non-convex) cycle-consistency constraint in Eq.~\eqref{eq: cycle}. This idea builds upon a \textit{virtual} universe shape, which can be thought of a shape that is never explicitly instantiated, i.e.~as opposed to a template shape we do not require a 3D mesh of the universe shape. Instead,  we merely assume that there is such a shape, so that for all points of the shapes in $\mathcal{S}$ there exists a corresponding (virtual) universe point. 
We denote the number of universe points as $d$.
For  $\Pi_{x} \in \{0, 1\}^{n_{x} \times d}$ being the matching from shape $\mathcal{X}$ to the universe shape, and $\Pi_{y}^{\top} \in \{0, 1\}^{d \times n_{y}}$ being the matching from the universe shape to shape $\mathcal{Y}$, this shape-to-universe representation allows to compute pairwise matchings as
    \begin{equation}
        \label{eq:shape-universe}
        \Pi_{xy} = \Pi_{x}\Pi_{y}^{\top}.
    \end{equation}

\section{Our unsupervised multi-shape matching method}
\begin{figure}[t!]
  \centering
  \includegraphics[width=\textwidth]{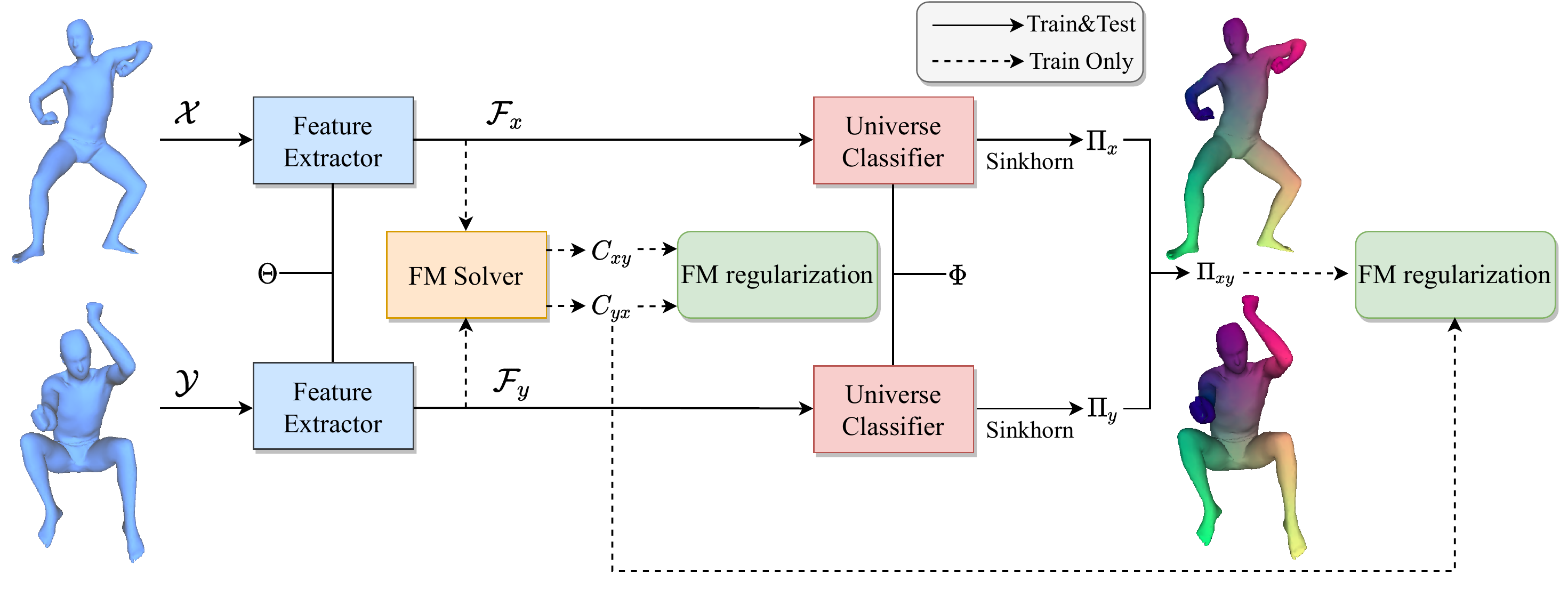}
  \caption{
  Overview of our unsupervised learning approach for cycle-consistent multi-shape matching. First, features $\mathcal{F}_{x}$ and $\mathcal{F}_{y}$ are extracted from the input shapes $\mathcal{X}$ and $\mathcal{Y}$.
  The feature descriptors are then used to compute the bidirectional functional maps $C_{xy}$ and $C_{yx}$ based on the FM solver. Subsequently, our novel universe classifier takes the feature descriptors as input and predicts a shape-to-universe matching ($\Pi_{x}$ and $\Pi_{y}$) for each shape. The pairwise matching $\Pi_{xy}$ is obtained by the composition of shape-to-universe point maps $\Pi_{x}\Pi_{y}^\top$. During training (solid and dashed lines) we utilise functional maps (FM) as regularisation, whereas at inference time no FM have to be computed (solid lines).
  }
  \label{fig:pipeline}
\end{figure}

Our novel unsupervised learning approach for cycle-consistent multi-shape matching is illustrated in Fig.~\ref{fig:pipeline}. Conceptually, our pipeline comprises of two main components that are trained in an end-to-end manner.

Analogous to other learning-based approaches~\cite{roufosse2019unsupervised,sharma2020weakly}, the first main component (blue in Fig.~\ref{fig:pipeline}) performs feature extraction. Given the source and target shapes $\mathcal{X}$ and $\mathcal{Y}$, a Siamese feature extraction network with (shared) trainable weights $\Theta$ extracts features $\mathcal{F}_{x}$ and $\mathcal{F}_{y}$ from the input shapes, respectively. Then, a (non-trainable but differentiable) FM solver (yellow in Fig.~\ref{fig:pipeline}) is applied to compute the bidirectional functional maps $C_{xy}$ and $C_{yx}$.
The second main component (red in Fig.~\ref{fig:pipeline}) is a Siamese universe classifier with shared weights $\Phi$. It takes features from the first part as input to predict the shape-to-universe matchings $\Pi_{x}$ and $\Pi_{y}$ for each shape. The pairwise matching $\Pi_{xy}$ is based on their composition, see Eq.~\eqref{eq:shape-universe}. To allow for an unsupervised end-to-end training of our architecture, we build upon functional map regularisation (green in Fig.~\ref{fig:pipeline}) described in Sec.~\ref{sec: fmaps}.
In the following we explain the individual parts in detail.

\subsection{Feature extractor}
The goal of the feature extraction module is to compute feature descriptors of  3D shapes that are suitable both for functional map computation and shape-to-universe matching prediction. 
Our feature extraction is applied in a Siamese manner, i.e.~the identical network is used for both $\mathcal{X}$ and $\mathcal{Y}$. The outputs of this module are point-wise feature descriptors for each shape, which we denote as $\mathcal{F}_{x}$ and $\mathcal{F}_{y}$ respectively. 

\subsection{Functional map solver}
\label{sec:fm_solver}
The FM solver aims to compute the bidirectional functional maps $C_{xy}$ and $C_{yx}$ based on the extracted feature descriptors $\mathcal{F}_{x}$ and $\mathcal{F}_{y}$ (see Sec.~\ref{sec: fmaps}). 
We use a regularised formulation to improve the robustness when  computing the optimal functional map~\cite{attaiki2021dpfm}, i.e.~we consider 
    \begin{equation}
        \label{eq:fmap_reg}
        C_{xy}=\underset{C}{\arg \min }\|C A_{x}-A_{y}\|_{F}^{2}+\lambda \sum_{i j} C_{i j}^{2} M_{i j},
    \end{equation}
where 
    \begin{equation}
        \label{eq:resolvent_mask}
        M_{i j}=\left(\frac{\Lambda_{y}(i)^{\gamma}}{\Lambda_{y}(i)^{2 \gamma}+1}-\frac{\Lambda_{x}(j)^{\gamma}}{\Lambda_{x}(j)^{2 \gamma}+1}\right)^{2}+\left(\frac{1}{\Lambda_{y}(i)^{2 \gamma}+1}-\frac{1}{\Lambda_{x}(j)^{2 \gamma}+1}\right)^{2}.
    \end{equation}
The regulariser can be viewed as an extension of Laplacian commutativity defined in Eq. \eqref{eq:lap} with well-justified theoretical foundation, see~\cite{ren2019structured} for details.

\subsection{Universe classifier}
The goal of the universe classifier module is to utilise the extracted feature descriptors $\mathcal{F}_{x}$ and $\mathcal{F}_{y}$ in order to predict the point-to-universe maps $\Pi_{x}$ and $\Pi_{y}$, respectively. Similarly to our feature extractor, the universe classifier is applied in a Siamese way. The output dimension of our universe classifier is equal to the number of points $d$ in the universe shape (see supp.~mat. for details how it is chosen).

{For our shape-to-universe matching representation, each point of a shape needs to be classified into exactly one universe class, and a universe class cannot be chosen multiple times (per shape). Mathematically, this means that the point-to-universe map $\Pi_x$ must be a partial permutation matrix as defined in Eq. \eqref{eq:pmap}, where in addition the rows of $\Pi_x$ sum up to one, i.e.
\begin{equation}\label{eq:perm}
    \Pi_{x} \in \left\{\Pi \in\{0,1\}^{n_{x} \times d}: \Pi \mathbf{1}_{d} = \mathbf{1}_{n_{x}}, \mathbf{1}_{n_{x}}^{\top} \Pi \leq \mathbf{1}_{d}^{\top}\right\}.
\end{equation}
We approximate these combinatorial constraints in terms of Sinkhorn normalisation~\cite{sinkhorn1967concerning,mena2018learning} to make the prediction differentiable.}
Sinkhorn normalisation iteratively normalises rows and columns of a matrix based on the softmax operator and a temperature parameter $\tau$ (see~\cite{mena2018learning}).

\subsection{Unsupervised loss}
Our unsupervised loss is composed of two main parts:

\paragraph{FM regularisation for feature extractor}
Following~\cite{roufosse2019unsupervised,sharma2020weakly}, we use functional map regularisation to compute unsupervised losses for optimised bidirectional functional maps $C_{xy}$ and $C_{yx}$. Specifically, we use the bijectivity loss $\mathcal{L}_{\mathrm{bij}}$ in Eq. \eqref{eq:bij}, the orthogonality loss $\mathcal{L}_{\mathrm{orth}}$ in Eq. \eqref{eq:orth}, and the Laplacian commutativity loss $\mathcal{L}_{\mathrm{lap}}$ in Eq. \eqref{eq:lap}  to regularise the functional maps. As such, the total loss for training the feature extractor can be expressed in the form
    \begin{equation}
        \label{eq:ft_unsup}
        \mathcal{L}_{\mathrm{ft}} = w_{\mathrm{bij}}\mathcal{L}_{\mathrm{bij}} + w_{\mathrm{orth}}\mathcal{L}_{\mathrm{orth}} + w_{\mathrm{lap}}\mathcal{L}_{\mathrm{lap}}.
    \end{equation}
In case of partial shape matching, the functional map from the complete shape to the partial shape becomes a slanted diagonal matrix~\cite{rodola2017partial}. We follow DPFM~\cite{attaiki2021dpfm} to regularise the predicted functional maps based on this observation. 
For $\mathcal{X}$ being the complete shape and $\mathcal{Y}$ being the partial shape, in the partial matching case the loss terms can be expressed as
\begin{equation}
     \mathcal{L}_{\mathrm{bij}}=\|C_{xy}C_{yx} - \mathbf{I}_{r}\|_{F}^{2}, \text{~and~}
     \mathcal{L}_{\mathrm{orth}}=\|C_{xy}C_{xy}^{\top} - \mathbf{I}_{r}\|_{F}^{2},
\end{equation}
where $\mathbf{I}_{r}$ is a diagonal matrix in which the first $r$ elements on the diagonal are equal to 1, and $r$ is the slope of the functional map estimated by the area ratio between two shapes, i.e.
    $r=\max \left\{i \mid \Lambda_{y}^{i}<\max \left(\Lambda_{x}\right)\right\}$.

\paragraph{FM regularisation for universe classifier} To train our universe classifier in an unsupervised manner, we build upon the relationship $\Phi_{x}C_{yx} \approx \Pi_{xy}\Phi_{y}$ between the functional map $C_{yx}$ and the point-wise map $\Pi_{xy}$, as explained in Eq. \eqref{eq:relation}.
In our case, the pairwise point map $\Pi_{xy}$ is obtained from the composition of the shape-to-universe point maps $\Pi_{xy}=\Pi_{x}\Pi_{y}^{\top}$. With that, the unsupervised loss for training the universe classifier can be expressed in the form
    \begin{equation}
        \label{eq:cls_unsup}
        \mathcal{L}_{\mathrm{cls}} = \|\Phi_{x}C_{yx}-\Pi_{xy}\Phi_{y}\|^{2}_{F}.
    \end{equation}
This loss is differentiable with respect to both the functional map $C_{yx}$ and the point map $\Pi_{xy}$, so that we can train our network in an end-to-end manner.
By doing so, the predicted functional map and the predicted point map are improved during training. 
Overall, we demonstrate that we are able to achieve better matching results even in comparison a network trained with ground truth correspondences (see Sec.~\ref{sec:ablation}).

The total loss combines the loss terms of the feature extractor and the universe classifier and has the form
    \begin{equation}
        \label{eq:total_unsup}
        \mathcal{L}_{\mathrm{total}} = \mathcal{L}_{\mathrm{ft}} + \lambda_{\mathrm{cls}} \mathcal{L}_{\mathrm{cls}}.
    \end{equation}
\subsection{Implementation details}
We implemented our method in PyTorch. Our feature extractor takes SHOT descriptors~\cite{salti2014shot} as inputs. We use DiffusionNet~\cite{sharp2020diffusionnet} as the network architecture for both our feature extractor and universe classifier. In terms of training, we use the Adam optimiser~\cite{kingma2015adam} with learning rate $1e^{-3}$ in all experiments. More details are provided in the supp.~mat.

\section{Experimental results}
\label{sec:results}
For our experimental evaluation, we consider complete shape matching, partial shape matching, as well as an ablation study that analyses the importance of individual components of our method. We note that in all experiments we train a single network for all shapes in a dataset (as opposed to a per-shape category network).

\begin{figure}[t]
\begin{minipage}{0.65\linewidth}
\resizebox{\linewidth}{!}{
 \scriptsize
        \centering
        \begin{tabular}{@{}lccccc@{}}
        \toprule
        \multicolumn{1}{c}{\textbf{Geodesic error ($\times$100)}}      & \multicolumn{1}{c}{\textbf{FAUST}}   & \multicolumn{1}{c}{\textbf{SCAPE}}  & \multicolumn{1}{c}{\textbf{F on S}}  & \multicolumn{1}{c}{\textbf{S on F}} & ${\bigtriangleup}$ \\ \midrule
        \multicolumn{6}{c}{Axiomatic Methods}  \\ 
        \multicolumn{1}{l}{BCICP \cite{ren2018continuous}}     & \multicolumn{1}{c}{6.4}  & \multicolumn{1}{c}{11}     & \multicolumn{1}{c}{-}    & - & \xmark   \\
        \multicolumn{1}{l}{ZOOMOUT \cite{melzi2019zoomout}}    & \multicolumn{1}{c}{6.1}  & \multicolumn{1}{c}{7.5}    & \multicolumn{1}{c}{-}       & - & \xmark  \\
        \multicolumn{1}{l}{Smooth Shells \cite{eisenberger2020smooth}} & \multicolumn{1}{c}{2.5} & \multicolumn{1}{c}{4.7}  & \multicolumn{1}{c}{-}  & - & \xmark   \\ \midrule
        \multicolumn{6}{c}{Supervised Methods} \\ 
        \multicolumn{1}{l}{FMNet \cite{litany2017deep}}       & \multicolumn{1}{c}{11}   & \multicolumn{1}{c}{17}     & \multicolumn{1}{c}{30}   & 33  & \xmark \\
        \rowcolor[HTML]{EFEFEF} 
        \multicolumn{1}{l}{+ \textit{pmf}} & \multicolumn{1}{c}{5.9}    & \multicolumn{1}{c}{6.3}       & \multicolumn{1}{c}{11}  & 14 & \xmark \\
        \multicolumn{1}{l}{3D-CODED \cite{groueix20183d}}     & \multicolumn{1}{c}{2.5}   & \multicolumn{1}{c}{31}    & \multicolumn{1}{c}{31}   & 33 & \cmark \\
        \multicolumn{1}{l}{GeomFmaps-KPC \cite{donati2020deep}} & \multicolumn{1}{c}{3.1}  & \multicolumn{1}{c}{4.4}  & \multicolumn{1}{c}{11} & 6.0 & \xmark \\
        \rowcolor[HTML]{EFEFEF} 
        \multicolumn{1}{l}{+ \textit{zo}}  & \multicolumn{1}{c}{1.9}          & \multicolumn{1}{c}{3.0}          & \multicolumn{1}{c}{9.2}          & 4.3  & \xmark \\
        \multicolumn{1}{l}{GeomFmaps-DFN \cite{sharp2020diffusionnet}}& \multicolumn{1}{c}{2.6}      & \multicolumn{1}{c}{3.0}   & \multicolumn{1}{c}{3.3} & 3.0 & \xmark  \\
        \rowcolor[HTML]{EFEFEF} 
        \multicolumn{1}{l}{+ \textit{zo}}  & \multicolumn{1}{c}{1.9}      & \multicolumn{1}{c}{2.4}  & \multicolumn{1}{c}{2.4}  & 1.9 & \xmark \\
        \multicolumn{1}{l}{HSN \cite{wiersma2020cnns}}    & \multicolumn{1}{c}{3.3}  & \multicolumn{1}{c}{3.5} & \multicolumn{1}{c}{25.4}  & 16.7 & \cmark  \\
        \multicolumn{1}{l}{ACSCNN \cite{li2020shape}}  & \multicolumn{1}{c}{2.7}     & \multicolumn{1}{c}{3.2}  & \multicolumn{1}{c}{8.4}  & 6.0 & \cmark   \\ \midrule
        \multicolumn{6}{c}{Unsupervised Methods}                                                                                           \\ 
        \multicolumn{1}{l}{SURFMNet \cite{roufosse2019unsupervised} }  & \multicolumn{1}{c}{15} & \multicolumn{1}{c}{12}  & \multicolumn{1}{c}{32} & 32 & \xmark \\
        \rowcolor[HTML]{EFEFEF} 
        \multicolumn{1}{l}{+ \textit{icp}} & \multicolumn{1}{c}{7.4}          & \multicolumn{1}{c}{6.1}          & \multicolumn{1}{c}{19}           & 23 & \xmark \\
        \multicolumn{1}{l}{Unsup FMNet \cite{halimi2019unsupervised}}  & \multicolumn{1}{c}{10} & \multicolumn{1}{c}{16}  & \multicolumn{1}{c}{29}  & 22 & \xmark \\
        \rowcolor[HTML]{EFEFEF} 
        \multicolumn{1}{l}{+ \textit{pmf}} & \multicolumn{1}{c}{5.7}          & \multicolumn{1}{c}{10}           & \multicolumn{1}{c}{12}           & 9.3  & \xmark \\
        \multicolumn{1}{l}{WSupFMNet \cite{sharma2020weakly}}  & \multicolumn{1}{c}{3.3}  & \multicolumn{1}{c}{7.3}  & \multicolumn{1}{c}{11.7}   & 6.2  & \xmark \\
        \rowcolor[HTML]{EFEFEF} 
        \multicolumn{1}{l}{+ \textit{zo}}  & \multicolumn{1}{c}{1.9}          & \multicolumn{1}{c}{4.9}          & \multicolumn{1}{c}{8.0}          & 4.3  & \xmark \\
        \multicolumn{1}{l}{Deep Shells \cite{eisenberger2020deep}}  & \multicolumn{1}{c}{1.7}  & \multicolumn{1}{c}{2.5} & \multicolumn{1}{c}{5.4}  & \textbf{2.7} & \xmark \\
        \multicolumn{1}{l}{Ours (\textit{fine-tune})} & \multicolumn{1}{c}{\textbf{1.5}}   & \multicolumn{1}{c}{\textbf{2.0}} & \multicolumn{1}{c}{7.3 ({\textbf{3.2}})}  & 8.6 (3.2) & \cmark \\ \hline
        \end{tabular}  
} 
\end{minipage}
\begin{minipage}{0.34\linewidth}
          \includegraphics[height=7.4cm]{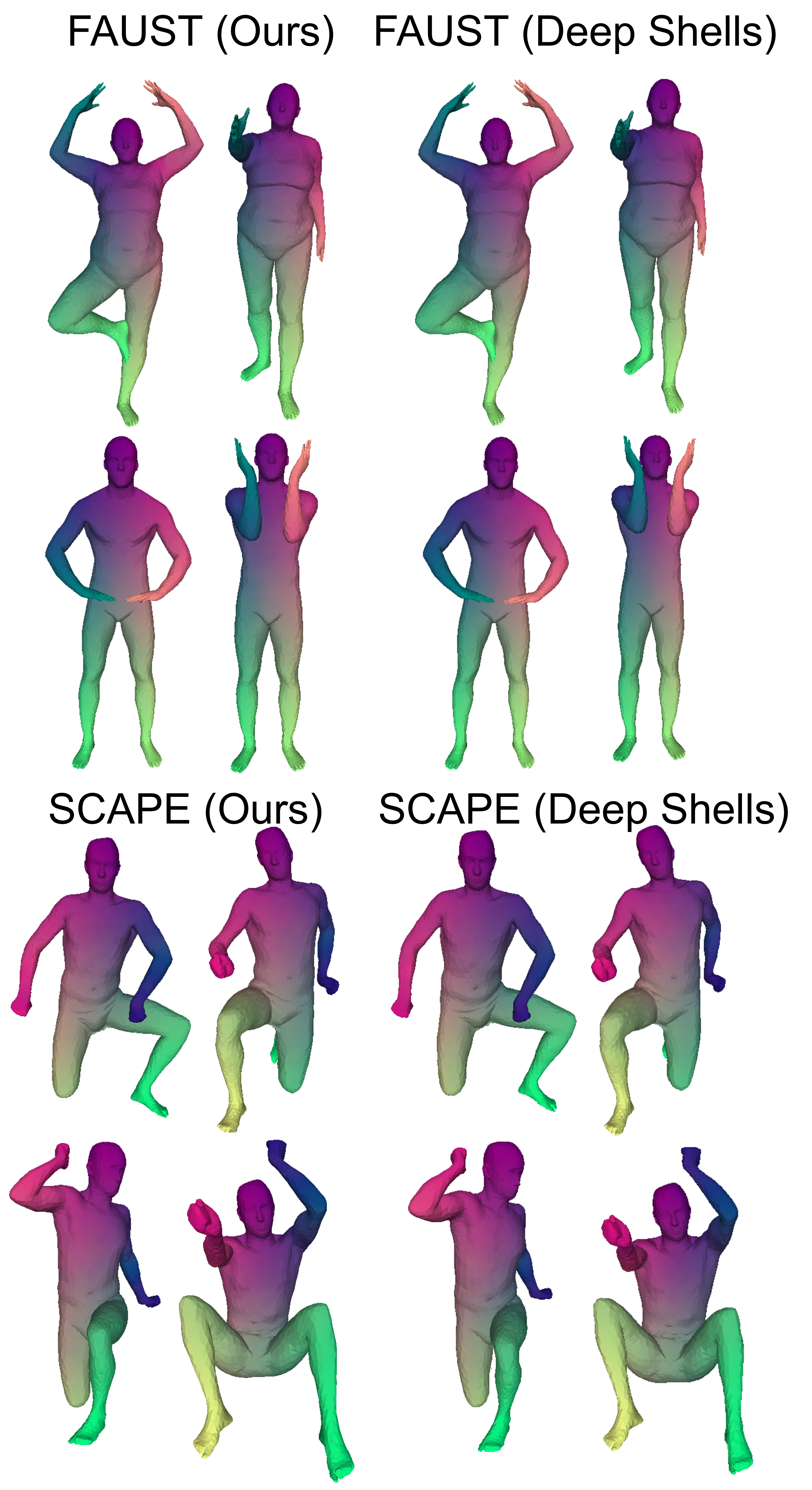}
\end{minipage}
    \caption{\textbf{Left:} Quantitative results on the FAUST and SCAPE datasets  in terms of mean geodesic errors ($\times$100). `F on S' stands for training on FAUST and testing on SCAPE datasets (`S on F' analogously). The rows in grey show refined results using the indicated post-processing procedure.  {Ours is the only unsupervised method that obtains cycle-consistent (${\bigtriangleup}$) multi-matchings.}
    \textbf{Right:} Qualitative multi-matching results using our method and Deep Shells. 
    {Although qualitatively both methods perform similarly, ours directly predicts shape correspondences without iterative refinement, thereby leading to a  faster inference, cf.~Fig.~\ref{fig:runtime}.
    }
    }
    \label{fig:fulleval}
\end{figure}

\subsection{Complete shape matching}
\label{sec:full_results}

\indent\paragraph{Datasets} To be consistent and comparable
with prior works, we evaluate our method on two standard benchmarks, FAUST~\cite{bogo2014faust} and SCAPE~\cite{anguelov2005scape} (for both, we use the more challenging remeshed versions from \cite{ren2018continuous}). The FAUST dataset contains 100 shapes consisting of 10 people, each in 10 poses. The SCAPE dataset comprises 71 different poses of the same person. Consistent with previous works, we split both datasets into training sets with 80 and 51 shapes, respectively, and test sets with 20 shapes.

\begin{figure}[t!]
\begin{center}
    \begin{subfigure}{.5\textwidth}
      \centering
      \includegraphics[width=\textwidth]{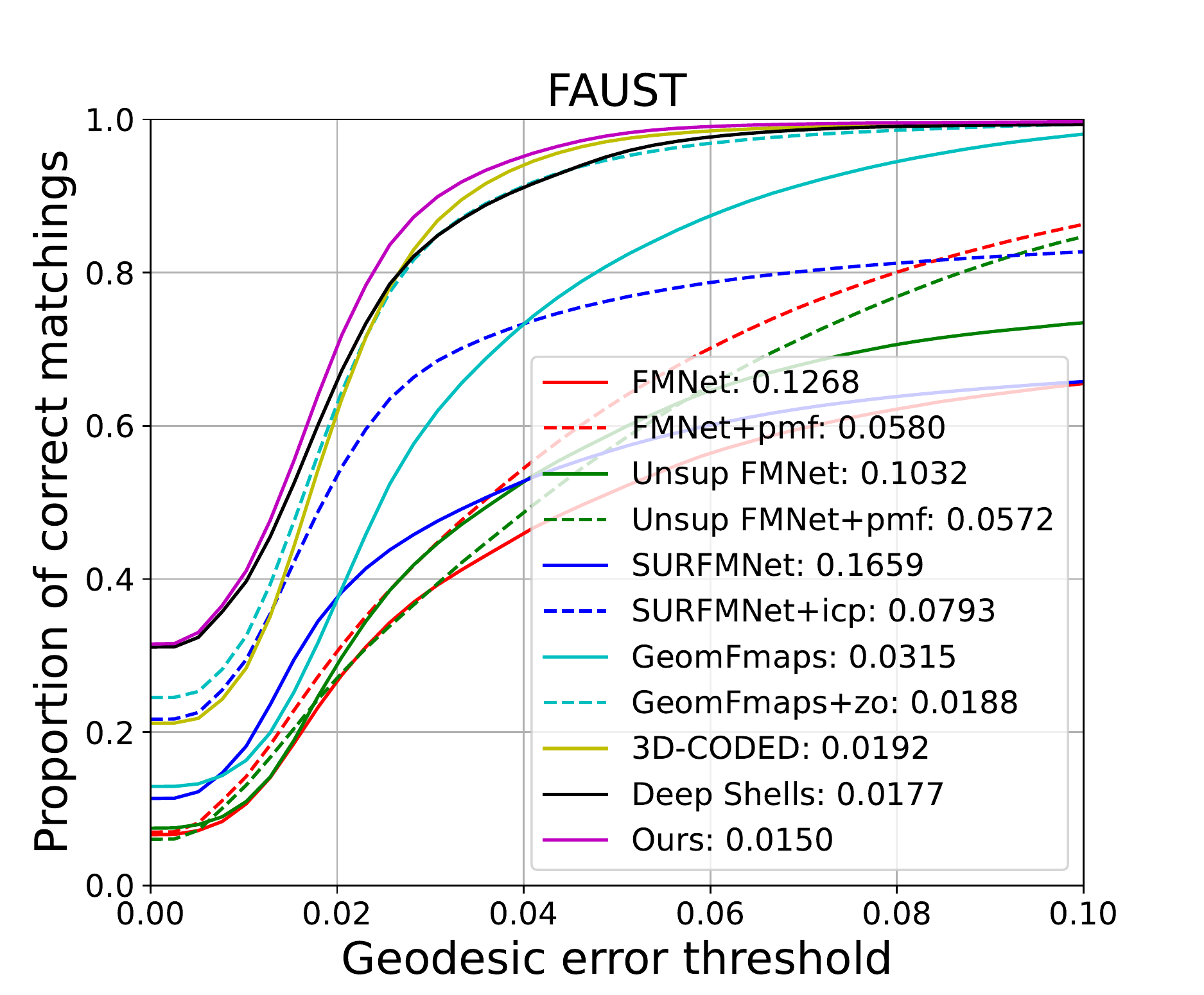}
      \label{fig:faust_pck}
    \end{subfigure}%
    \begin{subfigure}{.5\textwidth}
      \centering
      \includegraphics[width=\textwidth]{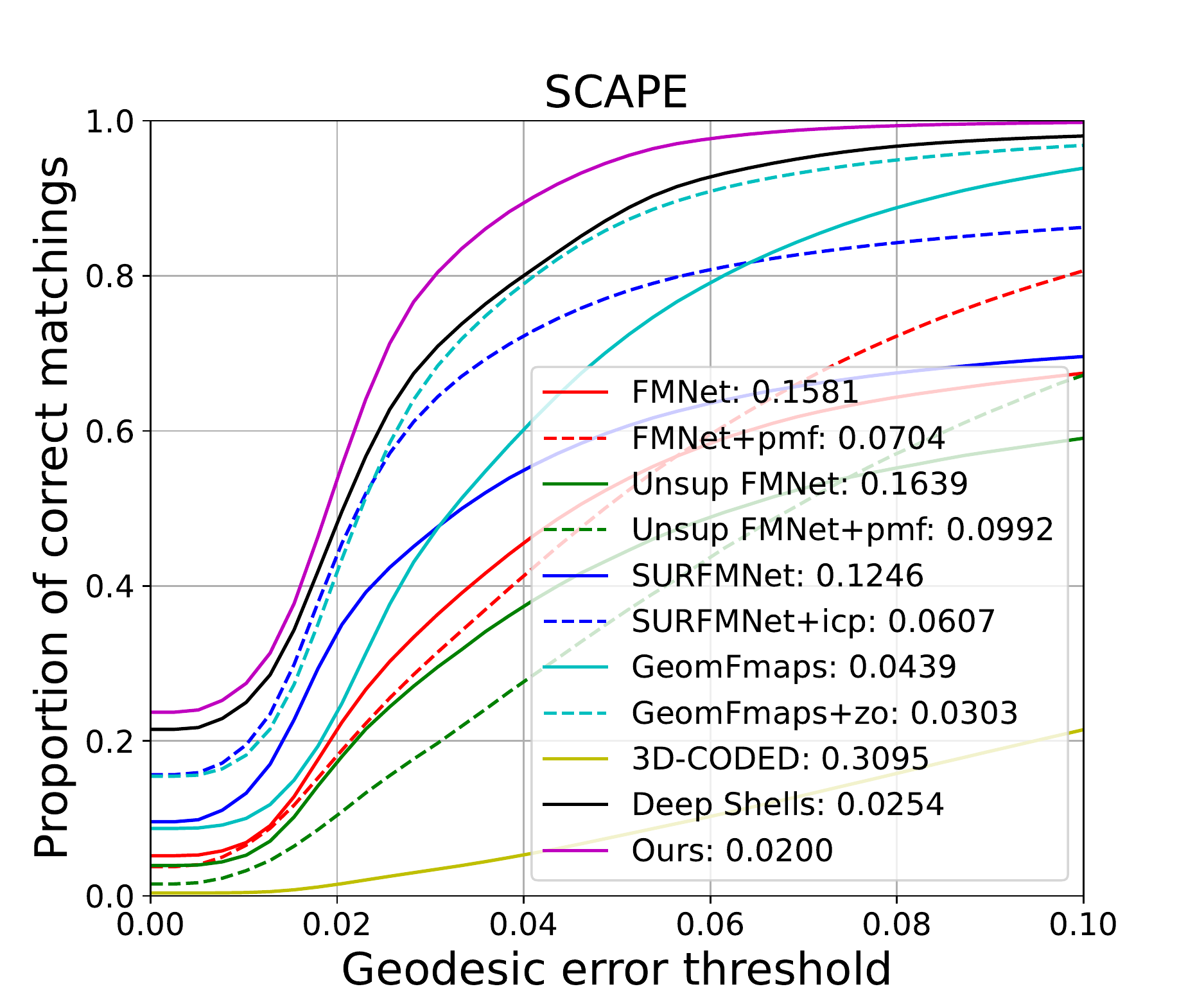}
      \label{fig:scape_pck}
    \end{subfigure}
    \caption{PCK curves for the FAUST and SCAPE dataset. Dashed lines indicate methods with refinement. Our method achieves the best scores on both datasets. 
    }
    \label{fig:pck}
\end{center}
\end{figure}

\begin{figure}[h!]
\begin{minipage}{0.55\linewidth}
 \scriptsize
        \centering
        \begin{tabular}{@{}lcccc@{}}
        \toprule
                  \textbf{Runtime (s)}             & \textbf{Inference} & \textbf{Refine} & \textbf{Total} \\ \midrule
        \multicolumn{4}{c}{Axiomatic Methods}                                                                          \\
        \multicolumn{1}{l}{BCICP~\cite{ren2018continuous}}             & 881.      & -      & 881.                        \\
        \multicolumn{1}{l}{ZoomOut~\cite{melzi2019zoomout}}            & 43.       & -      & 43.                        \\
        \multicolumn{1}{l}{Smooth Shells~\cite{eisenberger2020smooth}} & 125.      & -      & 125.                       \\ \midrule
        \multicolumn{4}{c}{Supervised Methods}                                                                         \\
        \multicolumn{1}{l}{FMNet~\cite{litany2017deep}}                & 5.1       & 223.   & 228.                      \\
        \multicolumn{1}{l}{3D-CODED~\cite{groueix20183d}}              & 725.      & -      & 725.                          \\
        \multicolumn{1}{l}{GeomFmaps-KPC~\cite{donati2020deep}}        & 1.9       & 35.    & 37.                         \\
        \multicolumn{1}{l}{GeomFmaps-DFN~\cite{sharp2020diffusionnet}} & 1.5       & 35.    & 37.                         \\ \midrule
        \multicolumn{4}{c}{Unsupervised Methods}                                                                       \\
        \multicolumn{1}{l}{SURFMNet~\cite{roufosse2019unsupervised}}   & 5.7       & 43.    & 49.                             \\
        \multicolumn{1}{l}{Unsup FMNet~\cite{halimi2019unsupervised}}  & 5.1       & 216.   & 221.                            \\
        \multicolumn{1}{l}{WSupFMNet~\cite{sharma2020weakly}}          & 2.1       & 35.    & 37.                             \\
        \multicolumn{1}{l}{Deep Shells~\cite{eisenberger2020deep}}     & 14.       & -      & 14.                             \\
        \multicolumn{1}{l}{Ours (\textit{fine-tune)}}                                         & \textbf{0.6} (5.0) & -      & \textbf{0.6} (5.0)                       \\ \bottomrule
        \end{tabular}
\end{minipage}
\begin{minipage}{0.45\linewidth}
    \centering
    \includegraphics[height=5.6cm]{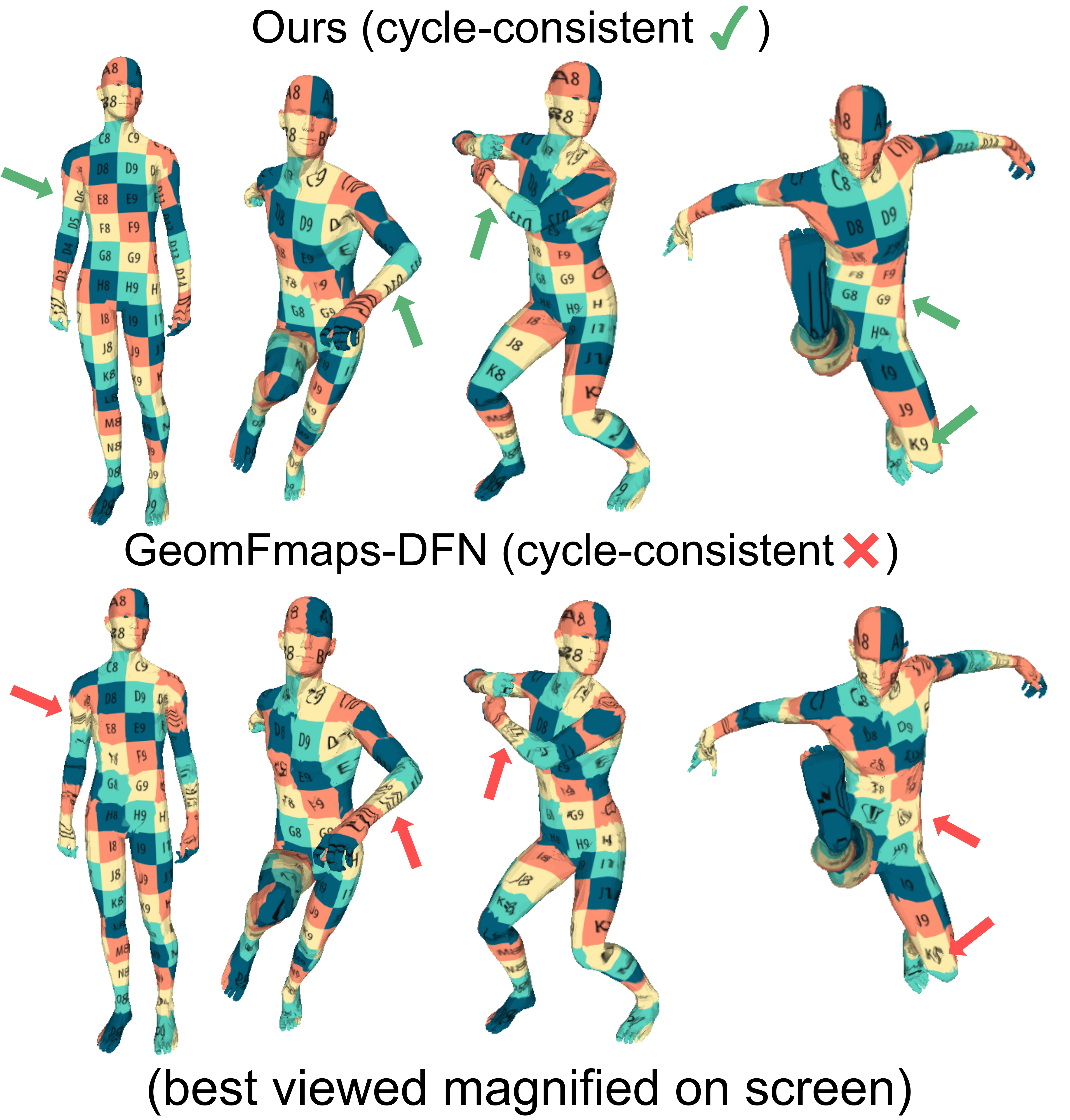}
\end{minipage}
    \caption{\textbf{Left:} Runtimes for the matchings in the experiments of Fig.~\ref{fig:fulleval}.  \textbf{Right:} Texture transfer using our method and the supervised GeomFmaps-DFN (erroneous matchings indicated by red arrows). 
    Cycle consistency ensures that our method consistently transfers textures.
    }
    \label{fig:runtime}
\end{figure}

\paragraph{Quantitative results}
For the evaluation we use the mean geodesic error defined in the Princeton benchmark protocol~\cite{kim2011blended}. We compare our method with state-of-the-art axiomatic, supervised and unsupervised methods, as shown in Fig.~\ref{fig:fulleval}. Our method outperforms the previous  state-of-the-art in most settings, even in comparison to the supervised methods. The last two columns in the table shown in  Fig.~\ref{fig:fulleval} (left) show generalisation results. Our method generalises better compared to previous unsupervised methods based on functional map regularisation~\cite{halimi2019unsupervised,roufosse2019unsupervised,sharma2020weakly}. In comparison to Deep Shells~\cite{eisenberger2020deep}, our method shows comparative results after fine-tuning our pipeline with the loss in Eq. \eqref{eq:total_unsup} for each shape pair individually (see details on fine-tuning in supp.~mat.). We plot the percentage of correct keypoints (PCK) curves for both datasets in Fig. \ref{fig:pck}, where it can be seen that our method achieves the best results in comparison to a wide range of methods.
Moreover, our method does not require post-processing techniques, such as ZOOMOUT~\cite{melzi2019zoomout} or PMF~\cite{vestner2017product}, which are often time-consuming, see Fig.~\ref{fig:runtime} for runtime comparisons. 

\subsection{Partial shape matching}
\paragraph{Datasets}
We evaluate our method in the setting of partial shape matching on the challenging SHREC'16 Partial Correspondence Benchmark \cite{cosmo2016shrec}. This dataset consists of 200 shapes with 8 categories (humans and animals). 
The dataset is divided into two subsets, namely CUTS (removal of a large part) with 120 pairs and HOLES (removal of many small parts) with 80 pairs. We follow the training procedure as in WSupFMNet \cite{sharma2020weakly}. 

\begin{figure}[t]
\begin{minipage}{0.4\linewidth}
\resizebox{\linewidth}{!}{
 \scriptsize
        \centering
        \begin{tabular}{@{}lccc@{}}
        \toprule
        \textbf{Geo.~error ($\times$100)}                   & \textbf{CUTS} & \textbf{HOLES} & ${\bigtriangleup}$ \\ \midrule
        \multicolumn{4}{c}{Axiomatic Methods}                         \\
        \multicolumn{1}{l}{FSP~\cite{litany2017fully}}        & 13.0 & 16.0 & \xmark \\ 
        \multicolumn{1}{l}{PFM~\cite{rodola2017partial}}    & 9.2 & 12.6 & \xmark \\ 
        \midrule
        \multicolumn{4}{c}{Unsupervised Methods}                      \\
        \multicolumn{1}{l}{WSupFMNet~\cite{sharma2020weakly}} & 15.0 & 13.0 & \xmark  \\
        \multicolumn{1}{l}{DPFM~\cite{attaiki2021dpfm}}       & 6.6  & 10.4 & \xmark \\
        \multicolumn{1}{l}{Ours}                                            & \textbf{5.5}  & \textbf{8.9} & \cmark  \\ \bottomrule
        \end{tabular}
} 
\end{minipage}
\begin{minipage}{0.6\linewidth}
    \centering
    \includegraphics[height=2.8cm]{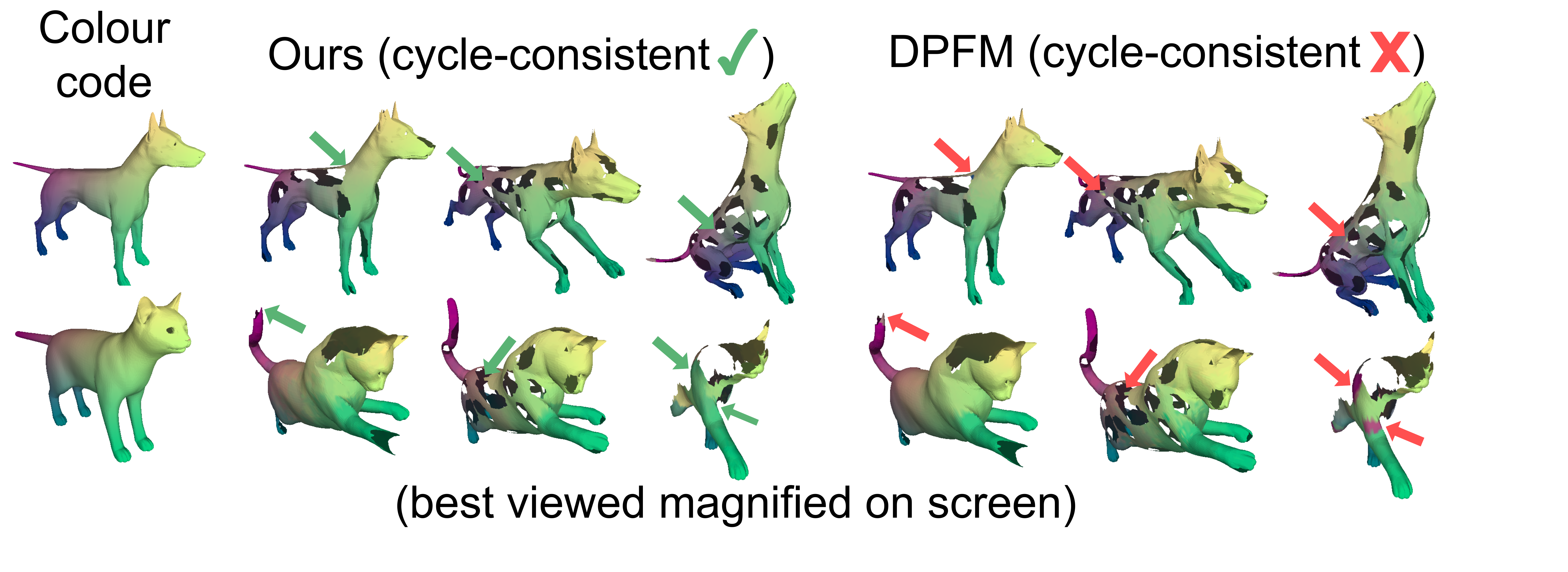}
\end{minipage}
    \caption{\textbf{Left:} Quantitative results on the CUTS and HOLES subsets of the SHREC'16 Partial Correspondence Benchmark. Reported are mean geodesic errors ($\times$100). {Ours is the only method that obtains cycle-consistent (${\bigtriangleup}$) multi-matchings.} \textbf{Right:} Qualitative  multi-matching results using our method and DPFM on the HOLES subset (erroneous matchings indicated by red arrows). 
    }
    \label{fig:partialeval}
\end{figure}

\paragraph{Quantitative results}
We compare our method with two axiomatic methods FPS \cite{litany2017fully}, PFM~\cite{rodola2017partial} and two unsupervised methods WSupFMNet~\cite{sharma2020weakly}, unsupervised DPFM~\cite{attaiki2021dpfm}. In Fig.~\ref{fig:partialeval} we show quantitative results (in terms of the mean geodesic error) and qualitative results. Our method outperforms previous axiomatic and unsupervised methods. The major difference in comparison to the unsupervised DPFM that predicts partial-to-complete {functional maps}, is that our method directly predicts shape-to-universe correspondences based on our universe classifier, so that ours leads to cycle-consistent multi-matchings. 

\subsection{Ablation study}
\label{sec:ablation}
\begin{figure}
\begin{minipage}{0.6\linewidth}
 \scriptsize
        \centering
        \begin{tabular}{@{}lcccc@{}}
        \toprule
        \multicolumn{1}{c}{\textbf{Geo.~error ($\times$100)}} & \textbf{FAUST} & \textbf{SCAPE} & \textbf{F on S} & \textbf{S on F} \\ \midrule
        \multicolumn{1}{l}{Classifier-free}                 & 2.1   & 3.8   & 17.4   & 22.9   \\
        \multicolumn{1}{l}{Feat. similarity}                & 2.1   & 3.7   & 10.6   & 13.9   \\
        \multicolumn{1}{l}{Supervised}                      & \textbf{1.4}   & 2.8   & 9.8    & 18.5   \\
        \multicolumn{1}{l}{Ours}                                                & 1.5   & \textbf{2.0}   & \textbf{7.3}    & \textbf{8.6}    \\ \bottomrule
        \end{tabular}
\end{minipage}
\begin{minipage}{0.4\linewidth}
    \centering
    \includegraphics[height=1.8cm]{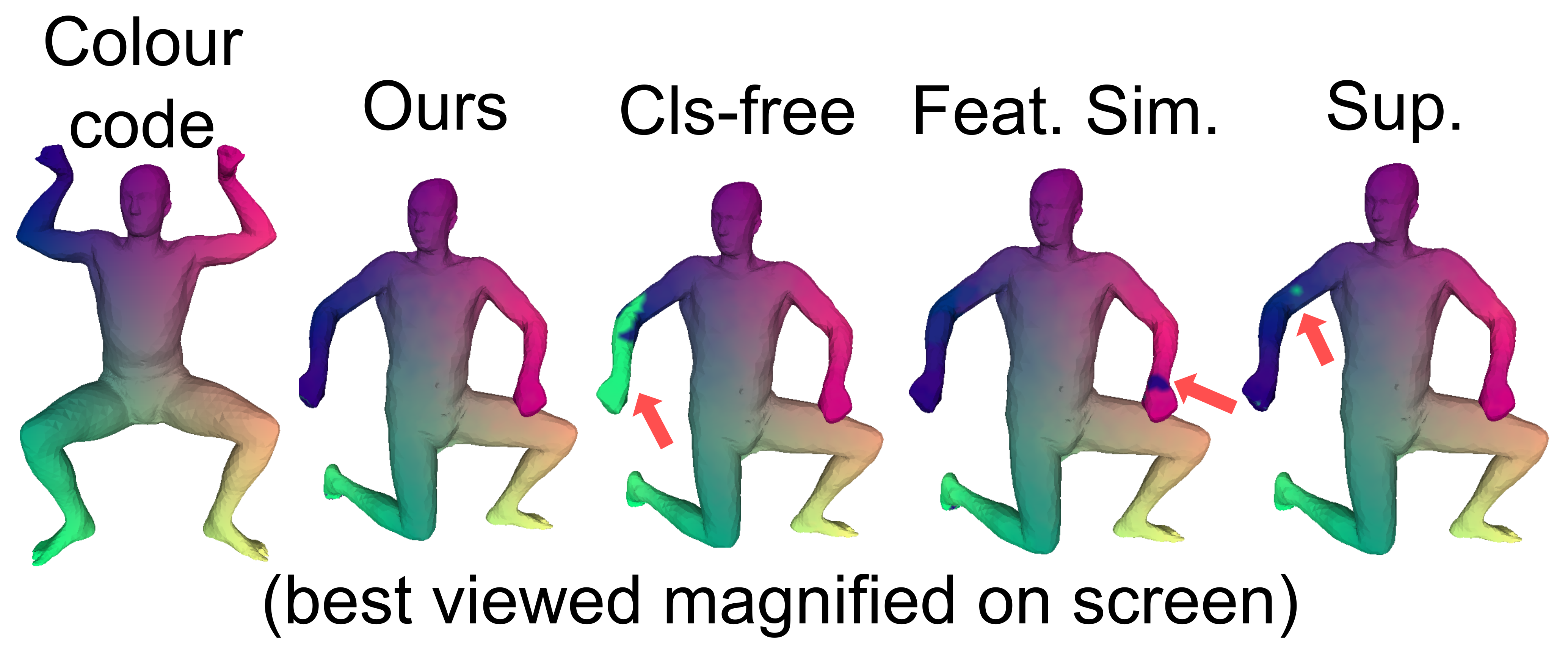}
\end{minipage}
    \caption{\textbf{Left:} Quantitative results of our ablation study on the FAUST and SCAPE datasets. \textbf{Right:} Qualitative results for the considered ablative experiments on the SCAPE dataset (erroneous matchings indicated by red arrows).
    }
    \label{fig:ablation}
\end{figure}

The goal of this section is to evaluate the importance of the individual components of our approach, including the universe classifier and our unsupervised loss.
For all ablation experiments, we consider the same experimental protocol as in
Sec.~\ref{sec:full_results}. We summarise the results for all ablative experiments and our full method in Fig.~\ref{fig:ablation}. We study a total of three different ablative settings that consider the removal of the universe classifier, or the modification of the loss. In the following we explain these in detail.

\paragraph{Classifier-free}
We remove the universe classifier in our pipeline and only train the feature extractor with the unsupervised loss $\mathcal{L}_{ft}$ defined in Eq. \eqref{eq:ft_unsup}. At test time, we convert the optimised functional maps to point maps using nearest neighbour search. In comparison to our complete method, the matching performance drops substantially, especially in the last two columns of Fig.~\ref{fig:ablation}, indicating that this results in poor generalisation  ability across datasets. Overall, this implies that our universe classifier is able to predict more accurate point maps compared to point maps converted from functional maps.

\paragraph{Feature similarity}
In the previous ablative experiment, we modify our network architecture by removing the classifier and change our unsupervised loss at the same time. Here, we  focus on the universe classifier only, i.e.~we remove it from our pipeline, while keeping our unsupervised loss. To this end, we construct a soft pairwise point map based on the feature similarity between two shapes with the help of Sinkhorn normalisation, similar to Deep Shells \cite{eisenberger2020deep}. By doing so, the pairwise point map can be expressed in the form $\Pi_{xy}(i, j) \propto \exp \left(-\frac{1}{\lambda}\left\|\mathcal{F}_{x}(i) - \mathcal{F}_{y}(j)\right\|_{2}^{2}\right)$. In comparison to classifier-free experiment, we observe that point maps based on such a feature similarity have similar performance on the intra-dataset experiments, while it significantly improves the generalisation ability across datasets. However, there is still a performance gap compared with our complete method.

\paragraph{Supervised training} The goal of this ablative experiment is to show the superiority of our unsupervised loss based on functional map regularisation compared to a supervised classification loss. In this experiment, we use the same network architecture as for our complete method. However, we replace our unsupervised loss defined in Eq. \eqref{eq:total_unsup} by a cross entropy loss between the predicted correspondences and the ground truth correspondences. In comparison to our complete method, we observe that the supervised alternative achieves better performance on FAUST dataset, but leads to a worse performance on other datasets, especially for the generalisation cases. We believe that the main reason is that the supervised approach is overfitting to the training data, but lacks the ability to generalise across datasets. 

\section{Limitations and future work}
Our work is the first unsupervised learning-based approach for finding cycle-consistent matchings between multiple 3D shapes, and additionally pushes the current state of the art in multi-shape matching. Yet, there are also some limitations that give rise to interesting future research questions.

For our approach it is not necessary that the universe shape is explicitly instantiated (e.g.~in the form of 3D mesh). 
However, we require to fix the maximum number of points $d$ of this (virtual) universe shape for training, since $d$ corresponds to the number of output classes that are predicted by our classifier. With that, a universe classifier trained with a fixed number of $d$ classes is not able to predict (unique) correspondences of shapes with more than $d$ vertices. 
The exploration of alternative formalisms that do not require to fix $d$ as a-priori is an interesting future direction.

Our universe classifier can be trained in an unsupervised manner both with datasets comprising of complete shapes only,  as well as with  mixed datasets (comprising of partial shapes and at least a single complete shape of the same category). 
As such, our current neural network does not allow for the {unsupervised} training with datasets that contain only partially observed shapes.
This is due to limitations for  partial-to-partial matchings that our approach inherits from the functional map framework. We  note that DPFM~\cite{attaiki2021dpfm} also shares this limitation -- in their case, the authors utilise a \emph{supervised} training strategy when considering the partial-to-partial matching case. Since DPFM uses different network architectures for partial-to-complete and partial-to-partial settings, it cannot be applied to different settings during training and inference. For example, it cannot be trained using partial-to-complete shape pairs and then be applied to predict partial-to-partial correspondences. In contrast,
our method is more flexible, as we use a single neural network architecture based on our universe classifier, where the shape-to-universe matching formalism naturally allows to represent complete, partial-to-complete as well as partial-to-partial matchings.

\section{Conclusion}
We introduce the first approach for the unsupervised training of a deep neural network for predicting cycle-consistent matchings across a collection of 3D shapes. Our approach builds upon the powerful functional map framework in order to allow for an unsupervised training. Yet, during inference we  directly predict point-wise matchings and do not require to compute functional maps, which has a positive impact on the runtime. The major strength of our approach is that it combines a unique set of favourable properties: our approach can be trained in an unsupervised manner, obtains cycle-consistent multi-matchings, does not rely on a template shape, and can handle partial matchings. Overall, due to the conceptual novelties and the demonstrated state-of-the-art performance on diverse shape matching benchmarks, we believe that our work advances the field of 3D shape matching.

\clearpage
%
%
\bibliographystyle{splncs04}
\bibliography{main}

\clearpage

\section{Supplementary Material}
In this supplementary document we first introduce the implementation details of our method. Subsequently, we provide details on the unsupervised loss for partial shape matching. Afterwards, we discuss our network fine-tuning. Eventually, we also show additional qualitative results of our method. 

\subsection{Implementation details}
We implemented our method in PyTorch. Our feature extractor takes 352-dimensional pre-computed SHOT descriptors~\cite{salti2014shot} as inputs. We use DiffusionNet~\cite{sharp2020diffusionnet} composed of 4 diffusion blocks with width 128 as the network architecture for both our feature extractor and universe classifier. In the context of the FM solver, we set $\lambda=100$ in Eq. \eqref{eq:fmap_reg} and $\gamma=0.5$ in Eq. \eqref{eq:resolvent_mask} for our partial shape matching (for complete shape matching we use $\lambda=0$). For the basis functions for functional maps computation, we choose the number to be 80 for the FAUST and SCAPE datasets for full shape matching. For partial shape matching, we choose the number to be 50 and 30 for the CUTS and HOLES subsets of the SHREC’16, respectively, to be consistent with DPFM~\cite{attaiki2021dpfm}. We apply Sinkhorn normalisation with the number of iterations equal to 10 and the temperature parameter $\tau$ equal to 0.2.

If a dataset provides ground truth correspondences based on a reference shape, we set the number of universe vertices to the number of vertices of the reference shape. Otherwise, we set the number of universe vertices to the largest number of vertices among the given shapes. For network training, we use $w_{\mathrm{bij}}=1.0, w_{\mathrm{orth}}=1.0, w_{\mathrm{lap}}=10^{-3}$ for $\mathcal{L}_{\mathrm{ft}}$ in Eq. \eqref{eq:ft_unsup} (for partial shape matching we use $w_{\mathrm{lap}}=0$, since in this case we already enforce Laplacian commutativity regularisation in our regularised FM solver).
The final loss is a linear combination of $\mathcal{L}_{\mathrm{ft}}$ and $\mathcal{L}_{\mathrm{cls}}$, where we set $\lambda_{\mathrm{cls}}=0.01$ for complete shape matching. The loss for the universe classifier $\mathcal{L}_{\mathrm{cls}}$ is slightly different from Eq. \eqref{eq:cls_unsup} for partial shape matching, for which we provide the details in Sec.~\ref{sec:unsup_loss}. We train our network with a batch size of 1 for all datasets. We use the ADAM optimiser with a learning rate of $10^{-3}$ for all experiments. The total number of training iterations for each dataset is 20000. During the first 4000 training iterations, when computing $\mathcal{L}_{\mathrm{cls}}$ defined in Eq. \eqref{eq:cls_unsup}, we detach the gradient for $C_{yx}$ and only regularise it based on its structural properties defined in $\mathcal{L}_{\mathrm{ft}}$. Afterwards, we will use the gradients for both $C_{yx}$ and $\Pi_{xy}$ to optimise our network. In this way, it can lead to faster convergence and better network performance.

\subsection{Unsupervised loss for partial shape matching}
\label{sec:unsup_loss}
In the context of partial-to-complete shape matching, we  can assume that the complete shape plays the role of the universe shape,
since it is guaranteed that each point in the partial shapes is in correspondence with exactly one point in the complete shape.
We modify the unsupervised loss for universe classifier based on it. For $\mathcal{X}$ being the complete shape and $\mathcal{Y}$ being the partial shape, the loss term can be expressed in the form
    \begin{equation}
        \mathcal{L}_{\mathrm{cls}} = \mathcal{L}_{\mathrm{ce}}^{\mathrm{smooth}}(\Pi_{x}, \mathbf{I}_{d}) + \mathcal{L}_{\mathrm{ce}}^{\mathrm{smooth}}(\Pi_{y}, \hat{\Pi}_{y}),
    \end{equation}
where $\mathbf{I}_{d}$ is the identity matrix of size $d$, $\hat{\Pi}_{y}$ is the partial-to-complete correspondences obtained by nearest neighbour search between $\Phi_{y}C_{xy}$ and $\Phi_{x}$, and $\mathcal{L}_{ce}^{\text{smooth}}$ is the cross entropy loss with label smoothing, where we set the smoothing factor equal to 0.1. The first term of the equation encourages the correspondences between the complete shape and the (virtual) universe shape to be identical, while the second term regularises the predicted partial-to-universe correspondence based on functional map regularisation. 
Similar to complete shape matching, the total unsupervised loss is a linear combination of $\mathcal{L}_{\mathrm{ft}}$ and $\mathcal{L}_{\mathrm{cls}}$, where we set $\lambda_{\mathrm{cls}} = 1.0$.

\subsection{Network fine-tuning}
We observe that the generalisation ability of our method across different datasets can be improved by network fine-tuning, as shown in Fig. \ref{fig:fulleval} (main paper). In order to achieve this, we first train our network on the training dataset in the ordinary way, and afterwards we use an unsupervised fine-tuning of the pre-trained network for the test dataset. Specifically, during fine-tuning, we  update the network weights for each shape pair independently. To this end, we use the same loss defined in Eq. \eqref{eq:total_unsup} to optimise the network with a fixed number of five  forward/backward passes (for each shape pair individually). The advantage of network fine-tuning compared to post-processing techniques is that it directly optimises the network itself, thus leading to better performance.

\subsection{Additional qualitative results}
\begin{figure}
    \centering
    \includegraphics[width=\textwidth]{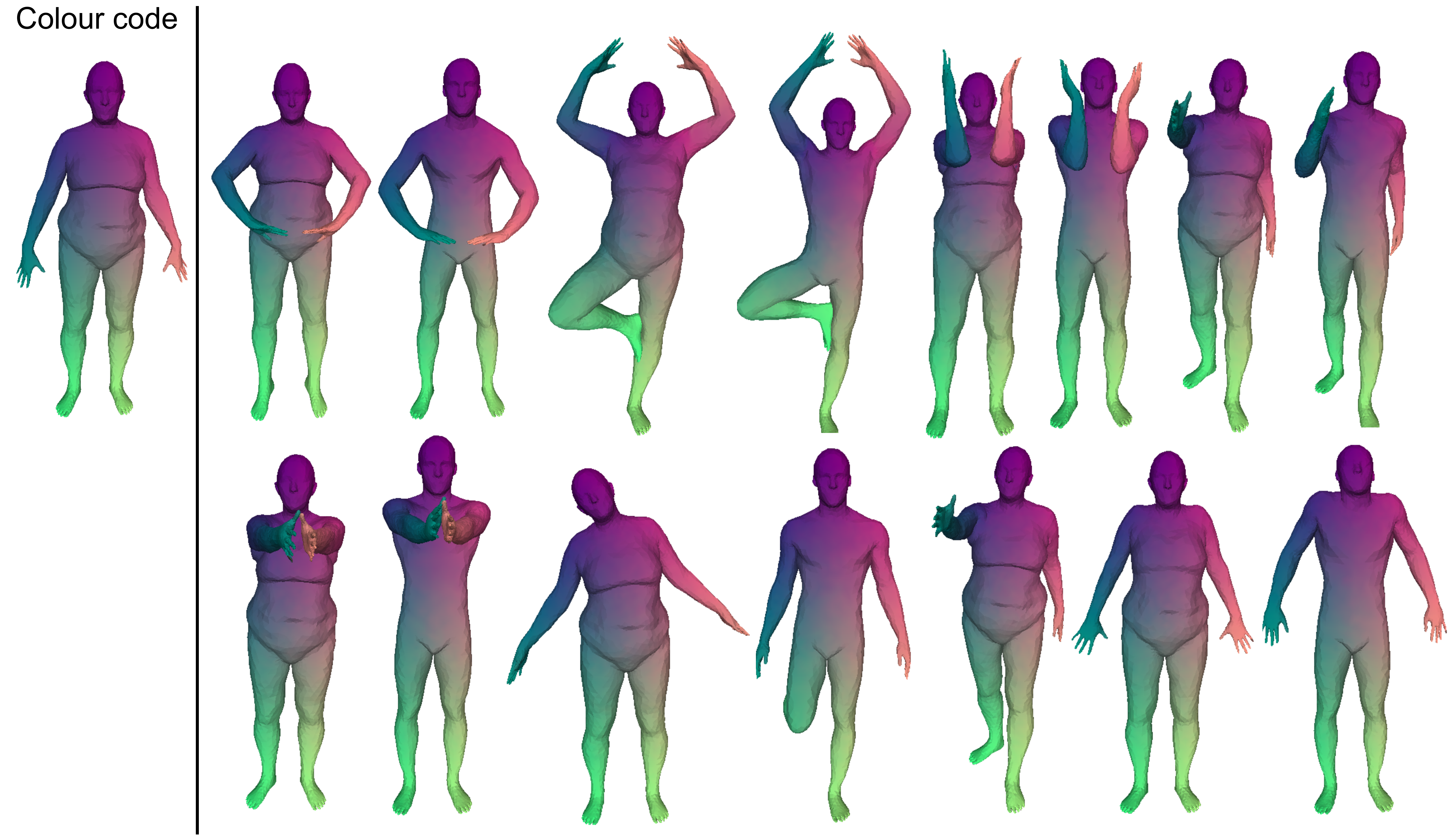}
    \caption{Qualitative multi-matching results using our method on the FAUST dataset.}
    \label{fig:faust}
\end{figure}

\begin{figure}
    \centering
    \includegraphics[width=\textwidth]{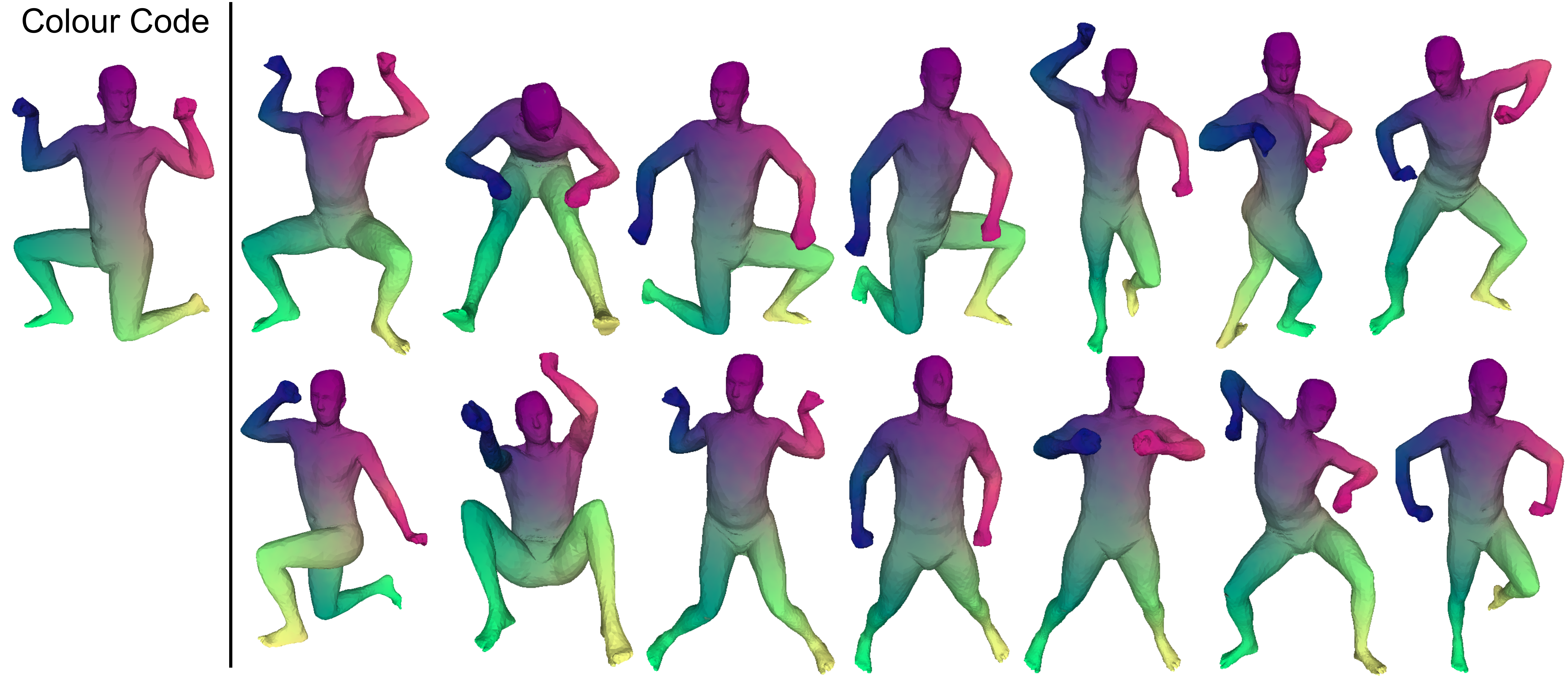}
    \caption{Qualitative multi-matching results using our method on the SCAPE dataset.}
    \label{fig:scape}
\end{figure}

\begin{figure}
    \centering
    \includegraphics[width=\textwidth]{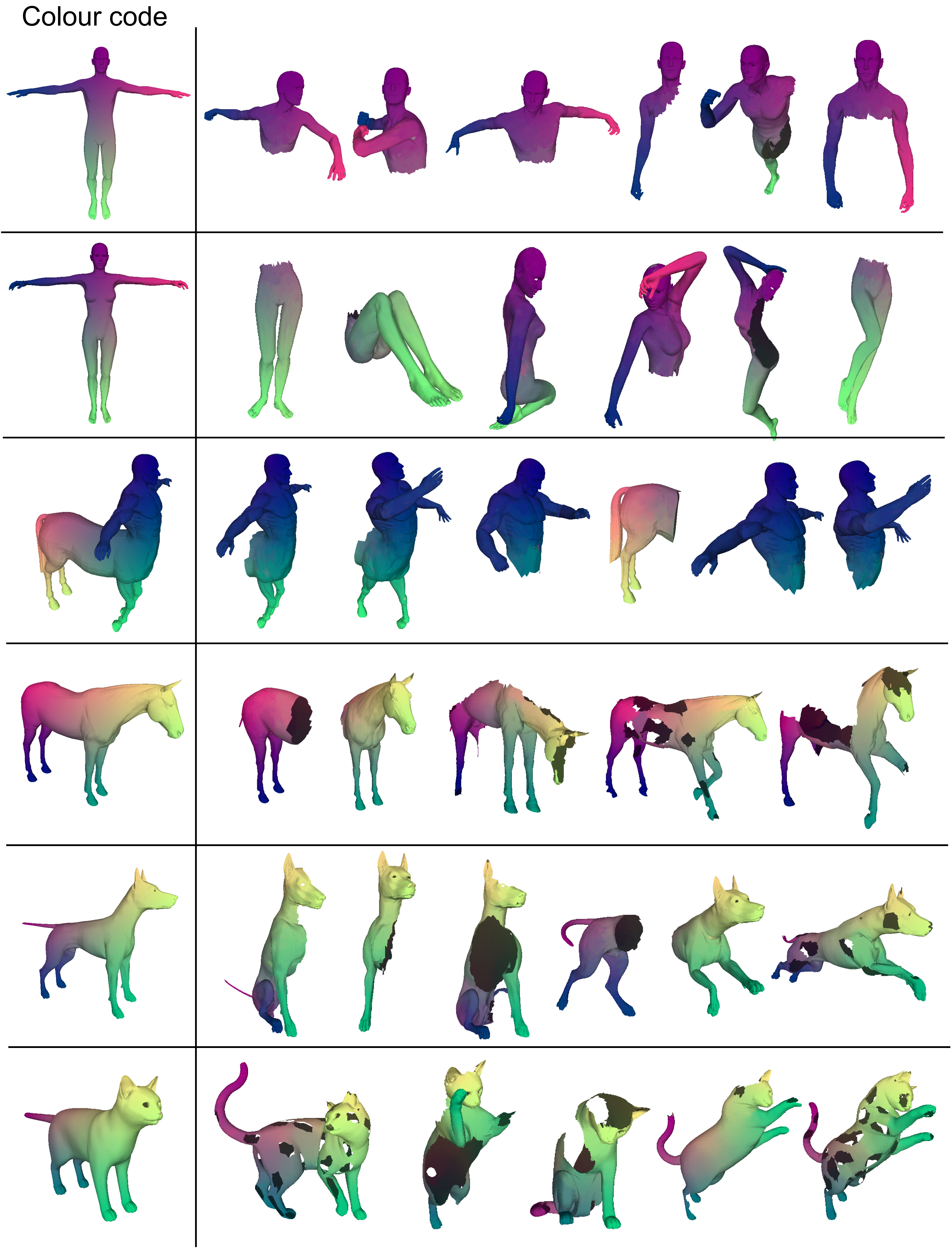}
    \caption{Qualitative partial-to-partial multi-matching results using our method on the SHREC'16 dataset. The full shape is shown merely for visualisation purposes (colour code).}
    \label{fig:shrec16}
\end{figure}

We show additional qualitative results on the FAUST dataset in Fig~\ref{fig:faust}, on the SCAPE dataset in Fig~\ref{fig:scape}, as well as on the SHREC'16 datset in Fig~\ref{fig:shrec16}. Our method predicts shape-to-universe correspondences for each shape to obtain cycle-consistent multi-shape matchings among a collection of shapes. 

\clearpage

\subsection{Inter-class shape matching}
We evaluate our method for the challenging inter-class multi-shape matching on the TOSCA dataset.
\begin{figure}[hbt!]
    \centering
    \includegraphics[width=\textwidth]{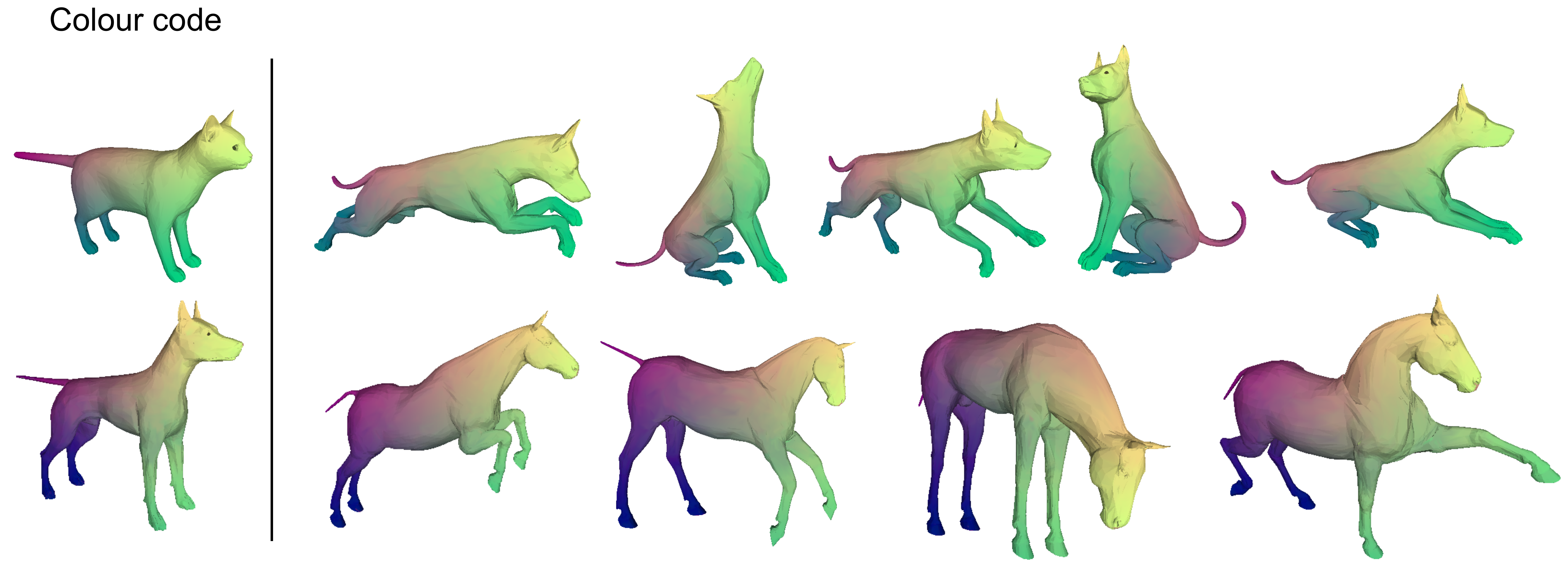}
    \caption{Qualitative inter-class multi-matching results using our method on the TOSCA dataset.}
    \label{fig:interclass}
\end{figure}

\subsection{Shape matching on SHREC'19 dataset}
We evaluate our method on the more challenging SHREC’19 dataset. Furthermore, we randomly remesh each shape to different resolution to evaluate the robustness of our method with respect to different meshings.
\begin{figure}[hbt!]
    \centering
    \includegraphics[width=\linewidth]{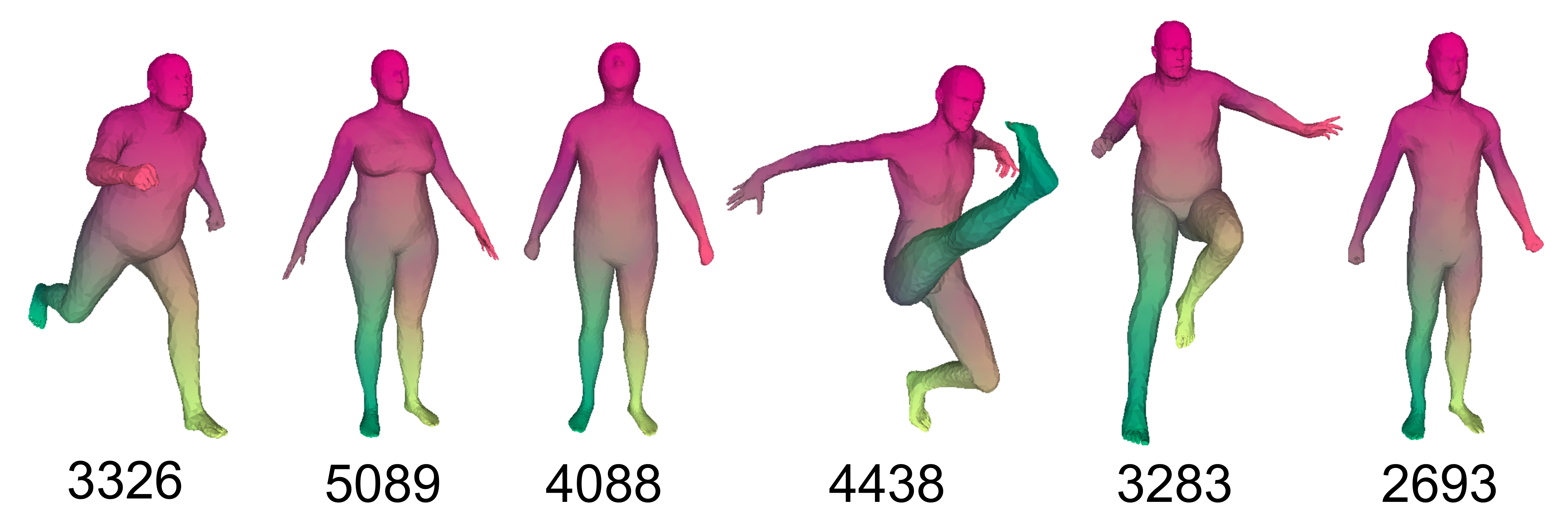}
    \caption{Qualitative shape matching results using our method on the SHREC'19 dataset with different resolution (numbers refer to \#vertices).}
    \label{fig:shrec19}
\end{figure}
\end{document}